\documentclass[conference]{IEEEtran}
\IEEEoverridecommandlockouts
\usepackage{cite}
\usepackage{graphicx}
\usepackage{amsmath,amssymb,amsfonts,amsthm}
\usepackage{graphicx}
\usepackage{textcomp}
\usepackage{comment,hyperref,float}
\usepackage{algorithm,algpseudocode}
\usepackage{multicol,multirow,lscape,hhline,ulem}
\usepackage[table,xcdraw]{xcolor}
\mathchardef\mhyphen="2D
\newtheorem{definition}{Definition}
\usepackage{subcaption}
\usepackage{mathabx}
\usepackage{bbold}
\usepackage{lineno}
\usepackage{hyperref}

\DeclareMathOperator*{\argmax}{arg\,max}

\begin{document}

\newcommand{\RFPc}[1]{\textcolor{blue}{#1}}
\newcommand{\RFPq}[1]{\textcolor{violet}{#1}}
\newcommand{\Change}[1]{\textcolor{olive}{#1}} 

\title{Crisp complexity of fuzzy classifiers}

\author{
\IEEEauthorblockN{1\textsuperscript{st} Raquel Fernandez-Peralta}
\IEEEauthorblockA{\textit{Mathematical Institute} \\
\textit{Slovak Academy of Sciences}\\
\textit{Bratislava, Slovakia}\\
raquel.fernandez@mat.savba.sk}
\and
\IEEEauthorblockN{2\textsuperscript{nd} Javier Fumanal-Idocin, 3\textsuperscript{rd} Javier Andreu-Perez}
\IEEEauthorblockA{\textit{School of Computer Science and Electronic Engineering} \\
\textit{University of Essex}\\
Colchester, United Kingdom \\
j.fumanal-idocin@essex.ac.uk}

}

\maketitle

\begin{abstract}
Rule-based systems are a very popular form of explainable AI, particularly in the fuzzy community, where fuzzy rules are widely used for control and classification problems. However, fuzzy rule-based classifiers struggle to reach bigger traction outside of fuzzy venues, because users sometimes do not know about fuzzy and because fuzzy partitions are not so easy to interpret in some situations. In this work, we propose a methodology to reduce fuzzy rule-based classifiers to crisp rule-based classifiers. We study different possible crisp descriptions and implement an algorithm to obtain them. Also, we analyze the complexity of the resulting crisp classifiers. We believe that our results can help both fuzzy and non-fuzzy practitioners understand better the way in which fuzzy rule bases partition the feature space and how easily one system can be translated to another and vice versa. Our complexity metric can also help to choose between different fuzzy classifiers based on what the equivalent crisp partitions look like.
\end{abstract}

\section{Introduction}

There is a growing need for transparency in modern machine learning systems, as they are deployed in many real-life scenarios where human and legal liabilities are present \cite{byrne2023good, arrieta2020explainable}. One of the most popular methods for explainable classification are rule-based and decision tree systems, which explicitly match patterns with outcomes \cite{timofeev2004classification, van2021evaluating, mendel2023explainable}, and this helps the user trust the decisions of the system.

Fuzzy rules have been one of the most popular methods in the fuzzy literature and research in this area remains active \cite{mendel2023explainable, fumanal2023artxai, salimi2022novel, fernandez2023subgroup}. Its principal lines of research are devoted to find the best rules possible and to obtain the better performance and explainability trade-offs \cite{mendel2023explainable, fumanal2024interpreting}.

Nonetheless, there exists a historical and ongoing reluctance surrounding the adoption of fuzzy logic techniques outside the fuzzy logic community \cite{Zadeh1995,Hullermeier2015}. Part of this mistrust relies in the fact that non-fuzzy users are not familiar with the theory of fuzzy logic and they may prefer crisp sets or methods purely based on probability theory and statistics. The interpretability of linguistic variables can also be debated, especially when the number of them is high or when the users are not particularly familiar with the vocabulary used \cite{mendel2021critical}.

However, fuzzy rules can also be understood as compact descriptions of underlying crisp decisions. For standard classification problems, each instance will be classified at the end of the process as a certain class, so the final output can be considered as crisp independently of the classifier used. Thus, a fuzzy classifier can also be translated into a crisp classifier. Based on this idea, we study in this paper the problem of constructing a crisp rule-based classifier equivalent to a fuzzy rule-based classifier, so that their decision boundaries are identical. We define two types of crisp partitions that behave exactly the same as the original classifier and we provide and implement an algorithm to explicitly construct them. The crisp description of the fuzzy classifier can help to disentangle the underlying complexity and it serves as a bridge for non-fuzzy users to help them by interpreting a more familiar output.  Further, the crisp description can also be used as an objective criterion for comparing the complexity and explainability of different fuzzy classifiers, even if the fuzzy partitions used are not the same.

The structure of the paper is as follows. First, in Section \ref{section:motivation} we describe the problem and we include the necessary background. In Section \ref{section:crisp_description} we depict two types of crisp descriptions of a fuzzy classifier and we provide the algorithms to construct them. In Section \ref{section:experiments} we build and compare the crisp descriptions of different classification algorithms and publicly available datasets. The paper ends in Section \ref{section:conclusions} with some conclusions and future work.

\section{Background}\label{section:motivation}

Let $\mathcal{D} = {(X_1,Y_1),\dots, (X_N,Y_N)}$ denote a data set with $N$ instances and $M$ features, where $X_i$ is the observed feature vector of the $i$-instance  with $j$-th entry $X_{i,j}$ and $Y_i$ being its discrete associated target. The target is a categorical variable with $C$ classes. We will denote the domain of each feature as $D_j \subseteq \mathbb{R}$ and the cartesian product of all the domains as $\mathcal{D} \subseteq \mathbb{R}^M$.

In general, the problem of multiclass classification consists of finding a function $f:\mathcal{D} \to \{1,\dots,C\}$ such that $f(X_i) = Y_i$ for as many instances as possible \cite{Bishop2006}. In this case, function $f$ is called a \textit{decision function} and it divides the input space into regions, one per class, such that each point in a certain region is assigned the corresponding class. The boundaries between decision regions are called \textit{decision boundaries}.

For comparing different classifications, there exists plenty of metrics in the literature \cite{Grandini2020}. However, since in this paper we are focused on explainability rather than performance, we only consider the accuracy score, which measures the percentage of correct predictions.

In rule-based classifiers, its decision function is constructed in terms of a collection of IF-THEN rules called the \textit{rule base} \cite{tan2019}. Every classification rule is usually expressed in the following manner:
$$
r : \text{IF }\text{Antecedent} \text{ THEN } Y \text{ IS } c,
$$
where ``Antecedent'' is a logical expression involving a subset of features and $c \in \{1,\dots,C\}$ is one of the possible classes.

Although all rule-based algorithms are described in terms of IF-THEN rules with the idea of a posterior interpretation of these rules by an expert,  distinct key elements can be chosen to design diverse classifiers: type of description language (crisp logic, fuzzy logic...); type of logical propositions (conjunction/disjunction of literals, disjunctive/conjunctive normal form...);  properties of the rule base (mutually exclusive rules, exhaustive rules, ordered rules, weighted rules...).

For instance, decision trees are considered to be one of the most simple rule-based classifiers since the paths from root to leaf represent the decision rules in which the classification is based. In this case, the rule base is mutually exclusive (each data point is assigned to exactly one class) and exhaustive (the entire feature space is covered).

With regard to the problem of interest in this paper, we will describe the problem of crisp and fuzzy rule-based classification in general, keeping the possible technicalities of each algorithm aside. Further, for simplifying the notation we consider the rules to be modeled as the conjunction of literals. For other kind of conditionals, our results can be easily adapted.

\subsection{Crisp rule-based classification}\label{subsec:cirsp_classification}

\noindent Let us consider a set $\mathcal{R}_{crisp}$ of $N_R = |\mathcal{R}_{crisp}|$ crisp rules, any rule $r \in \mathcal{R}_{crisp}$ with a maximum number of conditions in the antecedent $N_A$ is defined as follows:
\begin{equation*}
r : \text{IF } X \in S_r^1 \text{ AND } \dots \text{ AND } X \in S_r^{n_r} \text{ THEN } Y \text{ IS } Y_{r},
\end{equation*}
where $n_r \leq N_A$, $S_r^{j} \subseteq \mathcal{D}$ and $Y_r \in \{1,\dots,C\}$. Nonetheless, many classical crisp rule-based classification algorithms such as decision trees only consider crisp rules defined on hyperrectangles, i.e., 
\begin{multline}\label{eq:crisp_rule}
    r : \text{ IF } X_{r}^1 \in [\alpha_{r}^1,\beta_{r}^1] \text{ AND } \dots \text{ AND } X_{r}^{n_r} \in [\alpha_{r}^{n_r},\beta_{r}^{n_r}] \\ \text{ THEN } Y \text{ IS } Y_{r},
\end{multline}
where $X_{r}^j$ is one of the $M$ features with $[\alpha_{r}^j,\beta_{r}^j]$ laying inside the range of the corresponding variable. For the rest of this section, we will refer by a crisp rule in the sense of Eq. (\ref{eq:crisp_rule}). In this case, an example $x=(x_{1},\dots,x_{M})$ is classified as follows:

\begin{itemize}
	\item For each rule $r$ we compute the vector $(\mu_{r}^1(x_{r}^1),\dots,\mu_{r}^{n_r}(x_{r}^{n_r}))$ where  $\mu_{r}^j(x_{r}^j) = \mathbb{1}_{x_{r}^j \in [\alpha_{r}^j,\beta_{r}^j]}$, i.e., $\mu_{r}^j(x_{r}^j)=1$ if $x_{r}^j \in [\alpha_{r}^j,\beta_{r}^j]$ and 0 otherwise. Then, we denote $\displaystyle \mu_r(x) = \prod_{j=1}^{n_r}\mu_{r}^j(x_{r}^j)$. Notice that $\mu_r(x)=1$ if and only if $x$ fulfills all the conditions in the antecedent of the rule.
	\item We also compute $S_{r} \in \{0,1\}^C$ such as $S_{r,i} = 1$ if $i = Y_{r}$ and 0 otherwise.
	\item Then, we have two cases to do the classification depending on the characteristics of the set of rules:
	
	\begin{itemize}
		\item If the set of rules is mutually exclusive then
		$$f_{crisp}^{me}(x) = \argmax_{i\in \{1,\dots,C\}}  S_{\tilde{r},i},$$
        where $\tilde{r}=\argmax_{ r\in \mathcal{R}_{crisp}}\mu_r(x)$. This is the same as selecting which rule is satisfied by example $x$ and then assigning the example to the class marked by the consequent. 
		\item If the set of rules is not mutually exclusive then
		$$ f_{crisp}^{\neg me}(x) = \argmax_{i\in \{1,\dots,C\}} \sum_{s=1}^{N_R} S_{r_s,i}  \mu_r(x).$$
		This is equivalent to compute the number of rules that classify example $x$ to each class and taking the class that was most selected.
	\end{itemize}
\end{itemize}

\subsection{Fuzzy rule-based classification}\label{subsec:fuzzy_classification}

Let us first briefly recall the definition of fuzzy set and fuzzy linguistic variable.
\begin{definition}[\bf{\cite{Zadeh1965}}]\label{def:fuzzyset}
	Let $X$ be a classical set of objects, called the \emph{universe}. A \emph{fuzzy set} $A$ of $X$ is characterized by a function $\mu_A: X \to [0,1]$, which is called the \emph{membership function} of $A$. For all $x \in X$,  the value $\mu_A(x)$ indicates the \emph{degree of membership} of $x$ in the fuzzy set $A$.
\end{definition}
The \textit{core} and \textit{support} of a fuzzy set $A$ of $X$ with membership function $\mu_A$ are defined as $C_A= \{ x \in X \mid \mu_A(x)=1\}$ and $S_A= \{x \in X \mid \mu_A(x)>0\}$, respectively.
Various structures of membership functions for constructing fuzzy sets have been considered (triangular, trapezoidal, gaussian...). The selection of these functions depends on the specific problem and/or application \cite{Mitaim1996}. Nonetheless, trapezoidal membership functions are commonly chosen due to their simplicity and finite support.
\begin{definition}\label{def:trapezoidal_fs} 
Let $a,b,c,d \in \mathbb{R}$ with $a<b<c<d$, the corresponding trapezoidal membership function is defined by
				$$
		\mu_{trapezoidal}(x;a,b,c,d)= \left\{ \begin{array}{ll}
			0 &  \text{if }  x \leq a, \\[3pt]
			\frac{x-a}{b-a} & \text{if }  a < x \leq b,\\[3pt]
			1 & \text{if } b < x \leq c, \\[3pt]
			\frac{d-x}{d-c} &  \text{if }  c < x \leq d, \\[3pt]
			0 &  \text{if } x \geq d.
		\end{array}
		\right.
		$$
In this case, we will represent the fuzzy set by its parameters as follows $\langle a,b,c,d \rangle$. The core and support of a fuzzy set $A$ defined by a trapezoidal membership function is, $C_A = [b,c]$ and $S_A=(a,d)$, respectively.
\end{definition}
A fuzzy partition can be generally defined as a finite collection of fuzzy sets. 
 Further, a fuzzy linguistic variable is a variable whose values are expressed as words or phrases in either a natural or artificial language. In practice, for each linguistic variable involved in a concrete problem usually a fuzzy partition of its domain is considered.

Having said this, any fuzzy rule-based classifier includes a fuzzification step in which features are transformed into fuzzy linguistic variables. Specifically, if we consider the $j$-th feature then $\mu_{j,l}(X_{i,j})$ denotes the degree of truth of instance $i$ evaluated in the fuzzy set corresponding to the $l$-th label of the $j$-th feature. The number of labels of the $j$-th features is given by $l_{j}$. 

Now, let us consider a set $\mathcal{R}_{fuzzy}$ of $N_R = |\mathcal{R}_{fuzzy}|$ fuzzy rules with a maximum number of conditions in the antecedent $N_A$, any rule $r \in \mathcal{R}_{fuzzy}$ is defined as follows:
\begin{multline}
    r : \text{ IF } X_{r}^1 \text{ IS } L_{r}^1 \text{ AND } \dots \text{ AND } X_{r}^{n_r} \text{ IS } L_{r}^{n_r} \\ \text{ THEN } Y \text{ IS } S_{r},
\end{multline}
where $n_r \leq N_A$, $X_{r}^i$ is one of the $M$ features with $L_{r}^i$ being one of the linguistic labels of $X_r^i$  and $S_{r} \in \{0,1\}^C$ is a vector of scores assigned to each class. In this case, an example $x=(x_{1},\dots,x_{M})$ is classified as follows:

\begin{itemize}
	\item For each rule $r$ we compute the vector $(\mu_{r}^1(x_{r}^1),\dots,\mu_{r}^{n_r}(x_{r}^{n_r}))$ where $\mu_{r}^{j}(x_{r}^{j})$ is the corresponding membership value. Using the product as the T-norm, the truth degree of example $x$ for rule $r$ is given by
    $\displaystyle \mu_r(x) = \prod_{j=1}^{n_r}\mu_{r}^j(x_{r}^j) \in [0,1]$.
	\item Then, we may perform the classification in two different ways:
	\begin{itemize}
		\item \textit{Sufficient rules}:
		\begin{equation}\label{eq:fuzzy_sufficient}
            f_{fuzzy}^{suf}(x) = \argmax_{i\in \{1,\dots,C\}}  S_{\tilde{r},i},
            \end{equation}
        where $\tilde{r}=\argmax_{ r\in \mathcal{R}_{fuzzy}}\mu_r(x)$. This is equal to compute the truth degree of the rules, selecting the rule of the highest truth degree and then selecting the class that has a bigger score for that rule.
		\item \textit{Additive rules}:
		$$f_{fuzzy}^{ad}(x) = \argmax_{i\in \{1,\dots,C\}} \sum_{s=1}^{N_R} S_{r_s,i}  \mu_r(x).$$
		This is equivalent to compute the truth degree of the rules, multiply them by the score of each class, and then taking the class that has the highest value. 
	\end{itemize}
\end{itemize}

\section{Crisp equivalent of a fuzzy rule base}\label{section:crisp_description}

\subsection{Problem statement}
If we compare Sections \ref{subsec:cirsp_classification} and \ref{subsec:fuzzy_classification} there is a clear analogy between crisp and fuzzy classification. Indeed, the main difference is that in fuzzy classification examples do not generally fulfill a rule perfectly, but they have a truth degree between 0 and 1. That clearly affects the classification step, since fuzzy rules cannot be mutually exclusive by definition. Then, the problem of interest now arises almost naturally:
\begin{center}
    \textbf{Question:} Given a certain set of fuzzy rules $\mathcal{R}_{fuzzy}$, can we generate a set of crisp rules $\mathcal{R}_{crisp}$ such as $f_{crisp}(x) = f_{fuzzy}(x)$ for all $x \in \mathcal{D}$?
\end{center}
In the subsequent sections, we deeply study the problem and we provide an answer to this question. By simplicity, we will consider that the support of the fuzzy sets are intervals (like for trapezoidal fuzzy sets, see Definition \ref{def:trapezoidal_fs}), which is usually the case. Also, we will focus on sufficient rules, which are more easy to interpret because only one rule is used to classify each example.

\subsection{Crisp rule base defined on disjoint compatible regions}\label{subsec:disjoin_regions}
Let $\mathcal{R}_{fuzzy}$ be a set of fuzzy rules in the sense of Section \ref{subsec:fuzzy_classification}, then the support of rule $r$ is a subset of $\mathbb{R}^M$ that consists on the cartesian product of the support of all the fuzzy sets that appear in the antecedent and $\mathbb{R}$ for those features that are not involved in the rule. Specifically, if the features used in a rule $r$ are $\{X_r^1,\dots, X_r^{n_r}\}$ with the fuzzy sets $\{A_r^1,\dots, A_r^{n_r}\}$ corresponding to the labels $\{L_r^1,\dots, L_r^{n_r}\}$ then the support of $r$ is given by
$$
S_{r} = \bigtimes\limits_{i=1}^{M} S(A_i), \quad S(A_i) = \left\{
\begin{array}{ll}
      S_{A_i} & \text{if } A_i \in \{A_r^1,\dots, A_r^{n_r}\}, \\
      \mathbb{R} & \text{otherwise}.
\end{array} 
\right.
$$
Taking into account this notation, it is intuitive to think that the input space can be separated according to which rules are activated together, i.e., in those regions certain rules and only those rules jointly have support different than zero. Given a certain set of rules $R$ we define the regions of activation $B^{act}_R$ and joint activation $B^{jact}_R$ as follows
$$
B^{act}_R =  \bigcup_{r \in R} S_r, \quad B^{jact}_R =  \bigcap_{r \in R} S_r.
$$
By definition, $B^{act}_R$ (resp. $B^{jact}_R$) is the subergion of $\mathcal{D}$ in which at least 1 rule (resp. all the rules) in $R$ has non-zero support. Then,
\begin{equation}
B_R  = B_R^{jact} \setminus B_{R^c}^{act} =\left( \bigcap_{r \in R} S_r\right) \setminus \left( \bigcup_{r \not \in R} S_r\right),    
\end{equation}
is the region in which only the rules of $R$ are activated and we will call it \textit{compatible region}. If $B_R= \emptyset$ then we say that the rules in $R$ are incompatible, in the sense that they cannot be fired together. If $B_R \not = \emptyset$ the rules are compatible and the data inside this region is classified according to the truth degrees of the rules involved and their scores. The union of all these regions for all the possible subsets of $\mathcal{R}_{fuzzy}$ defines the region of the universe covered by the algorithm, i.e., 
$$\mathcal{R}_i = \{R \subseteq \mathcal{R}_{fuzzy} \mid |R| = i\}, \quad i \in \{1,\dots,N_R\},$$ 
$$\Omega_{\mathcal{R}_{fuzzy}} = \bigcup_{i=1}^{N_R} \bigcup_{R \in \mathcal{R}_i} \left( \bigcap_{r \in R} S_r\right) \setminus \left( \bigcup_{r \not \in R} S_r\right).$$
Ideally, $\Omega_{\mathcal{R}_{fuzzy}} = \mathcal{D}$, but depending on the algorithm it may happen that not the whole space is covered. Also, notice that by construction all the regions are disjoint.

Having said this, we can design an iterative algorithm to compute the non-empty compatible regions $B_R$ and then compute all the corresponding crisp rules according to Eq. (\ref{eq:fuzzy_sufficient}). Further, it is clear that if $B_R^{jact} = \emptyset$ then for any $\tilde{R} \subseteq R$ we have that $B_{\tilde{R}}^{jact} = \emptyset$  and the algorithm can have a similar structure than the well-known Apriori \cite{Agrawal1994} to optimize the search. The pseudo-code of the algorithm can be found in Algorithm \ref{alg:1}.

\begin{algorithm}[htp]
\caption{Disjoint regions crisp rule miner}\label{alg:1}
\begin{algorithmic}
\Require A set of fuzzy rules $\mathcal{R}_{fuzzy}$.
\Ensure A set of crisp rules $\mathcal{R}_{crisp}$ such that $f_{crisp} = f_{fuzzy}$.
\State $\mathcal{R}_{crisp} \leftarrow \emptyset$.
\For{$\{r\} \in \mathcal{R}_1$}
\If{$B_{\{r\}} \not = \emptyset$}
\State $\mathcal{R}_{crisp} \leftarrow$ ``If $X \in S_r$ THEN $Y$ IS $Y_r$''. 
\EndIf
\EndFor
\State $C_1,L_1 = \mathcal{R}_1$.
\State $k=2$.
\While{$L_{k-1} \not = \emptyset$}
\State $C_k \leftarrow $ \textsc{combine\_regions}($L_{k-1}$).
\For{$R \in C_k$}
\If{$B_R^{jact} \not = \emptyset$}
\State $L_k \leftarrow R$.
\EndIf
\State $B_R = \left( \bigcap_{r \in R} S_r\right) \setminus \left( \bigcup_{r \not \in R} S_r\right)$.
\If{$B_R \not = \emptyset$}
\For{$r \in R$}
\State $\mathcal{R}_{crisp} \leftarrow$ ``If $X \in B_{R}$ AND $\mu_{r}(X) \geq \underset{\underset{S_{r} \not = S_{\tilde{r}}}{r \not = \tilde{r}}}{\max} \mu_{\tilde{r}}(X)$ THEN $Y$ is $Y_{r}$''.
\EndFor
\EndIf
\EndFor
\State $k \leftarrow k+1$.
\EndWhile
\State \textbf{return} $\mathcal{R}_{crisp}$.
\end{algorithmic}
\end{algorithm}
\subsection{Crisp rule base defined on disjoint hyperrectangles}\label{subsec:disjoin_hyperrectangles}

Although Algorithm \ref{alg:1} computes a set of crisp rules that is minimal in the sense that only takes into account non-empty disjoint regions of the feature space in which rules are activated together, the rules are evaluated in the regions $B_R$ which are not hyperrectangles in general. Indeed, the union or difference of hyperrectangles is not usually a hyperrectangle. This may be seen as an important drawback for some users, since crisp rules defined in complex domains are not so easy to interpret. For this reason, in this section we provide an alternative approach in which the crisp rules obtained are defined on disjoint hyperrectangles.

Let again $\mathcal{R}_{fuzzy}$ be a set of fuzzy rules, we define the set of all the supports of the rules as $\mathcal{S}_{\mathcal{R}_{fuzzy}} = \{S_r \mid r \in \mathcal{R}_{fuzzy}\}$. Since we have considered only fuzzy sets whose support is an interval, each $S_r$ is a hyperrectangle. Then, $\mathcal{S}_{\mathcal{R}_{fuzzy}}$ is a set of possibly overlapping hyperrectangles that we can transform in another set $\mathcal{H}_{\mathcal{R}_{fuzzy}}$ of disjoint hyperrectangles that cover the exact same space. 

Then, for each $H \in \mathcal{H}_{\mathcal{R}_{fuzzy}}$ we will have a set of rules $R_H$ that have non-zero support in $H$, and we can proceed similarly than Section \ref{subsec:disjoin_hyperrectangles} to compute the corresponding crisp rules by taking into account Eq. (\ref{eq:fuzzy_sufficient}).

\subsection{Crisp rule bases as a complexity measure} 
If we contemplate the case in which all the possible compatible regions $B_R$ are non-empty, we can obtain the following upper bound for the number of crisp rules 
\begin{equation}
|\mathcal{R}_{crisp}| \leq \sum_{i=1}^{N_R} \binom{N_R}{i} \cdot i = N_R \cdot 2^{N_R-1},
\label{eq:upper_bound}
\end{equation}
which would grow quite fast to intractability (see Table \ref{tab:upper_bounds}).  However, in practice the number of crisp rules is expected to be much lower, as we will see in Section \ref{section:experiments}.
\begin{table}[h]
\caption{Values of the upper bound in Eq. (\ref{eq:upper_bound}) for different total number of fuzzy rules $N_R$.}
\centering
\begin{tabular}{c|c|c|c|c|c|c|c|c|c}
\textbf{$N_R$} & 2 & 3  & 4  & 5  & 6   & 7   & 8    & 9    & 10   \\ \hline
(\ref{eq:upper_bound}) & 4 & 12 & 32 & 80 & 192 & 448 & 1024 & 2304 & 5120
\end{tabular}
\label{tab:upper_bounds}
\end{table}

In fact, it may happen that the size of the crisp equivalents of two fuzzy rule bases completely differ even if the number of fuzzy rules is very similar. Then, the number of crisp rules can be used as a measure of complexity to compare different fuzzy rule bases. The lower the size of the crisp rule base the simpler is the feature space splitted and therefore the easier is to interpret. Notice that this metric is agnostic to the chosen fuzzy partitions, which is a problematic when comparing classifiers with different fuzzifications. If we want to measure how complex is the classifier with respect to the worst-case scenario, the upper bound (\ref{eq:upper_bound}) can be used to compute the degree of complexity:
 \begin{equation}\label{eq:complexity}
 \frac{|\mathcal{R}_{crisp}|}{N_R \cdot 2^{{NR-1}}}.
 \end{equation}

\section{Experiments}\label{section:experiments}
For testing our methods, different fuzzy rule-based classifiers have been implemented with the exFuzzy library \cite{fumanal2024}, which uses genetic fine tuning to fit the classifiers to the data.  This implementation allows custom selection of different parameters like granularity of fuzzy partitions, maximum number of conditions in the antecedent or maximum number of rules. In this section, if not specified otherwise we have used the parameters by default and fuzzy partitions are generated according to the data distribution. To match the modeling considered in this paper, we have considered unweighted rules.

\begin{figure}[htp]
    \centering
    \begin{subfigure}[b]{0.24\textwidth}
        \centering
        \includegraphics[width=\textwidth]{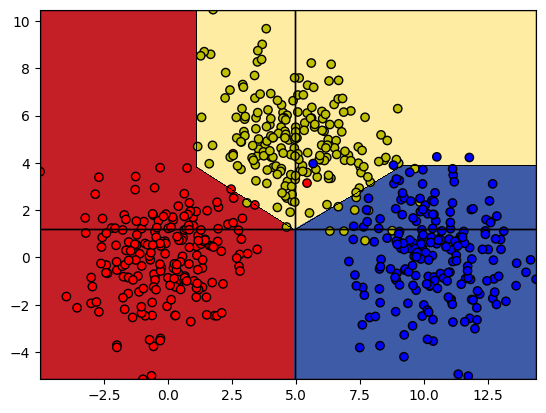}
        \caption{Fuzzy Classifier.}
        \label{fig:fuzzy_boundaries}
    \end{subfigure}
    \begin{subfigure}[b]{0.24\textwidth}
        \centering
        \includegraphics[width=\textwidth]{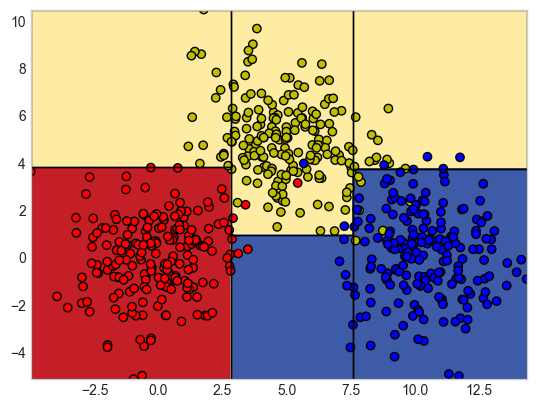}
        \caption{Decision tree.}
        \label{fig:crisp_boundaries}
    \end{subfigure}

    \caption{Plot of the decision boundary of two classifiers, where $0=\text{Red}$, $1=\text{Yellow}$ and $2=\text{Blue}$.}
    \label{fig:boundaries}
\end{figure}

First of all, we provide an illustrative example of the procedure of obtaining a crisp rule base given a fuzzy one. For this purpose, we have considered a 3-class synthetic dataset with only two numeric features, $X_1$ and $X_2$ (see the scatter plot in any of the subfigures of Figure \ref{fig:boundaries}). The two features have been fuzzified as two fuzzy linguistic variables with three linguistic terms as follows:
\begin{align*}
    \mu_{Low}^1 &: \langle -\infty,-\infty,1.12,4.96\rangle, \nonumber \\
    \mu_{Medium}^1 &: \langle 1.12,3.04,7.09,9.23\rangle, \nonumber \\
    \mu_{High}^1 &: \langle 4.96, 9.23, +\infty,+\infty\rangle, \nonumber \\
    \mu_{Low}^2 &: \langle -\infty, -5.15, -0.52, 1.19\rangle, \nonumber \\
    \mu_{Medium}^2 &: \langle -0.52, 0.34, 2.54, 3.88 \rangle, \nonumber \\
    \mu_{High}^2 &: \langle 1.19, 3.88, +\infty, +\infty\rangle. \tag{FP}\label{eq:fp} 
\end{align*}
Then, from the genetic fuzzy classifier we obtain the following fuzzy rule base:
\begin{align}
    r_1 &: \text{IF } X_1 \text{ IS Low THEN } Y \text{ IS } 0. \nonumber \\
    r_2 &: \text{IF } X_2 \text{ IS High THEN } Y \text{ IS } 1. \nonumber \\
    r_3 &: \text{IF } X_1 \text{ IS High THEN } Y \text{ IS } 2. \tag{FRB}\label{eq:frb}
\end{align}
In this case, there is one fuzzy rule per class and each rule only involves a certain variable with a corresponding fuzzy set. The support of each rule is
$$S_{r_1} = (-\infty,4.96) \times \mathbb{R}, \quad S_{r_2} = \mathbb{R} \times (1.19,+\infty)$$
$$S_{r_3} = (4.96,+\infty) \times \mathbb{R},$$
and the only non-empty compatible regions are 
\begin{eqnarray*}
    B_{r_1} &=& (-\infty,4.96) \times (\infty,1.19], \\
    B_{r_1,r_2} &=& (-\infty,4.96) \times (1.19,+\infty), \\
    B_{r_2,r_3} &=& (4.96,+\infty) \times (1.19,+\infty), \\
    B_{r_3} &=& (4.96,+\infty) \times (\infty,1.19],
\end{eqnarray*}
which separate $\mathbb{R}^2$ in 4 disjoint squares, so in this case it is not necessary to split these regions because they already are hyperrectangles. Since the three rules point to different classes, in regions $B_{r_1,r_2}$ and $B_{r_2,r_3}$ the final classification takes into account the comparison between 
$\mu_{Low}^1$ and $\mu_{High}^2$ in $B_{r_1,r_2}$ and $\mu_{High}^1$ and $\mu_{High}^2$ in $B_{r_2,r_3}$. Then, it is straightforward that with Algorithm \ref{alg:1} we obtain the following crisp rule base:
\begin{align}
    \tilde{r}_1 &: \text{IF } (X_1,X_2)  \in (-\infty,4.96) \times (-\infty,1.19] \text{ THEN } Y \text{ IS } 0. \nonumber \\
    \tilde{r}_2 &: \text{IF } (X_1,X_2)  \in (-\infty,4.96) \times (1.19,+\infty)  \nonumber\\
& \text{ AND } \mu_{Low}^1(X_1) \geq \mu_{High}^2(X_2) \text{ THEN } Y \text{ IS } 0. \nonumber \\
    \tilde{r}_3 &: \text{IF } (X_1,X_2)  \in (-\infty,4.96) \times (1.19,+\infty)  \nonumber\\
    & \text{ AND } \mu_{Low}^1(X_1) \leq \mu_{High}^2(X_2) \text{ THEN } Y \text{ IS } 1. \nonumber \\
    \tilde{r}_4 &: \text{IF } (X_1,X_2)  \in (4.96,+\infty) \times (1.19,+\infty) \nonumber \\
    & \text{ AND } \mu_{High}^1(X_1) \leq \mu_{High}^2(X_2) \text{ THEN } Y \text{ IS } 1. \nonumber \\
    \tilde{r}_5 &: \text{IF } (X_1,X_2)  \in (4.96,+\infty) \times (1.19,+\infty) \nonumber \\
    & \text{ AND } \mu_{High}^1(X_1) \geq \mu_{High}^2(X_2) \text{ THEN } Y \text{ IS } 2. \nonumber \\
    \tilde{r}_6 &: \text{IF } (X_1,X_2)  \in (4.96,+\infty) \times (-\infty,1.19] \text{ THEN } Y \text{ IS } 2. \tag{CRB}\label{eq:crb}
\end{align}

In Figure \ref{fig:fuzzy_boundaries} we can see the region delimited by the 3 fuzzy rules and the 6 crisp rules (since by definition the decision boundary of the two rule bases is the same, we have highlighted the boundaries of the compatible regions also in black to help the visualization). For comparison purposes, we have trained a decision tree fixing the maximum number of leaves to 6 (see Figure \ref{fig:crisp_boundaries}). In this case, the fuzzy rule base (\ref{eq:frb}) successfully describes in a compact manner the three regions of the space clearly dominated by each class. This is possible because the presence of the fuzzy sets allows a more complex support for the rules than hyperrectangles. Indeed, in Figure \ref{fig:crisp_boundaries} we can see how the decision tree is forced to create 3 disjoint regions to separate class 1 from the other two classes. From the linguistic representation of (\ref{eq:frb}) we can interpret that class 0 is related with low values of $X_1$, class 1 with high values of $X_2$ and class 2 with high values of $X_1$. However, from this representation only it is not clear how the three rules exactly interrelate, which may result confusing for the expert. From the crisp description (\ref{eq:crb}) not only we untangle how classification is performed, but also we can derive a more profound interpretation of how the classifier works. Indeed, classification can be explained in plain words as follows: if $X_2$ is not at all High (outside the support of the fuzzy set), then the instance is classified to 0 if $X_1$ is Low with any degree and to 1 otherwise (i.e. it is high by the construction of the fuzzy partition since the fuzzy sets of High and Low do not overlap). On other hand, if $X_2$ is High and $X_1$ is Low with any degree, then instance is classified to 0 if $X_1$ is more Low than $X_2$ is High and 1 otherwise. Finally, the only remaining situation is that $X_2$ and $X_1$ are High with any degree. In this case, the instance is classified as 1 if $X_2$ is more High than $X_1$ is High and 2 otherwise.


\begin{table*}[htp]
\centering
\caption{Results of the exfuzzy algorithm per dataset for different configurations of total number of rules and number of fuzzy sets in fuzzy partitions and other crisp rule-based algorithms. For exfuzzy, in each cell it appears a tuple with the following values: number of fuzzy rules, number of crisp rules, complexity, and crisp rules defined on hyperrectangles (top) and test accuracy (bottom). For the other algorithms only the number of crisp rules is included at the top. Also, under the name of each dataset there is a summary of the following specifications: (samples, features, classes).}\label{table:datasets}
\begin{tabular}{c|c|c|c|c|c|c|c|}
\cline{2-8}
                                    & \textbf{exfuzzy 3-10} & \textbf{exfuzzy 3-20} & \textbf{exfuzzy 5-10} & \textbf{exfuzzy 5-20}  & \textbf{Tree} & \textbf{RIPPER-k} & \textbf{PART}  \\ \hline
                                    
\multicolumn{1}{|c|}{\begin{tabular}[c]{@{}c@{}}\textbf{appendicitis}\\ (106,7,2)\end{tabular}} &  \begin{tabular}[c]{@{}c@{}}(6,69,0.36,501)\\ 0.89\end{tabular} & \begin{tabular}[c]{@{}c@{}}(10,571,0.11,8192)\\ 0.91\end{tabular}  & \begin{tabular}[c]{@{}c@{}}(7,54,0.12,499)\\ 0.86\end{tabular} &  \begin{tabular}[c]{@{}c@{}}(8,44,0.04,890)\\ 0.91\end{tabular}  & \begin{tabular}[c]{@{}c@{}} 13\\ 0.86 \end{tabular} & \begin{tabular}[c]{@{}c@{}} 2\\ 0.86 \end{tabular} & \begin{tabular}[c]{@{}c@{}} 2\\ 0.83 \end{tabular} \\ \hline

\multicolumn{1}{|c|}{\begin{tabular}[c]{@{}c@{}}\textbf{iris}\\ (365,4,3)\end{tabular}} &   \begin{tabular}[c]{@{}c@{}}(5,16,0.2,121)\\ 0.98\end{tabular}    &  \begin{tabular}[c]{@{}c@{}}(4,19,0.59,43)\\ 0.76\end{tabular} & \begin{tabular}[c]{@{}c@{}}(4,4,0.13,15)\\ 0.96\end{tabular} &  \begin{tabular}[c]{@{}c@{}}(7,26,0.05,182)\\ 0.86\end{tabular}  & \begin{tabular}[c]{@{}c@{}} 6\\ 0.94 \end{tabular} & \begin{tabular}[c]{@{}c@{}} 3\\ 0.94 \end{tabular} & \begin{tabular}[c]{@{}c@{}} 6\\ 0.94 \end{tabular} \\ \hline

\multicolumn{1}{|c|}{\begin{tabular}[c]{@{}c@{}}\textbf{magic}\\ (19020,10,2)\end{tabular}} & \begin{tabular}[c]{@{}c@{}}(6,143,0.74,2197)\\ 0.81\end{tabular}  &  \begin{tabular}[c]{@{}c@{}}(8,270,0.26,14996)\\ 0.76\end{tabular} & \begin{tabular}[c]{@{}c@{}}(9,208,0.09,25104)\\ 0.69\end{tabular} &     \begin{tabular}[c]{@{}c@{}}(9,83,0.04,12140)\\ 0.73\end{tabular}   & \begin{tabular}[c]{@{}c@{}} 1336\\ 0.81 \end{tabular} & \begin{tabular}[c]{@{}c@{}} 27\\ 0.85 \end{tabular} & \begin{tabular}[c]{@{}c@{}} 40\\ 0.84 \end{tabular} \\ \hline

\multicolumn{1}{|c|}{\begin{tabular}[c]{@{}c@{}}\textbf{vehicle}\\ (846,18,4)\end{tabular}} & \begin{tabular}[c]{@{}c@{}}(8,686,0.7,104102)\\ 0.44\end{tabular}  &  \begin{tabular}[c]{@{}c@{}}(10,1024,0.2,372944)\\ 0.55\end{tabular}  &     \begin{tabular}[c]{@{}c@{}}(6,86,0.45,123450)\\ 0.37\end{tabular}  & \begin{tabular}[c]{@{}c@{}}(10,820,0.16,340288)\\ 0.45 \end{tabular} &  \begin{tabular}[c]{@{}c@{}} 94\\ 0.68 \end{tabular}  & \begin{tabular}[c]{@{}c@{}} 16\\ 0.68 \end{tabular} & \begin{tabular}[c]{@{}c@{}} 27\\ 0.725 \end{tabular}    \\ \hline

\multicolumn{1}{|c|}{\begin{tabular}[c]{@{}c@{}}\textbf{australian}\\ (690,14,2)\end{tabular}} & \begin{tabular}[c]{@{}c@{}}(5,63,0.79,1399)\\ 0.82\end{tabular}  &  \begin{tabular}[c]{@{}c@{}}(7,45,0.1,1828)\\ 0.8\end{tabular} & \begin{tabular}[c]{@{}c@{}}(3,4,0.33,48)\\ 0.86\end{tabular} &     \begin{tabular}[c]{@{}c@{}}(7,30,0.07,8512)\\ 0.85\end{tabular}    &  \begin{tabular}[c]{@{}c@{}} 58\\ 0.82 \end{tabular} & \begin{tabular}[c]{@{}c@{}} 4\\ 0.86 \end{tabular} & \begin{tabular}[c]{@{}c@{}} 28\\ 0.85 \end{tabular}  \\  \hline
\end{tabular}
\end{table*}


Now, we study the number of crisp rules for different fuzzy classifiers. In particular, we have considered the configurations of 3 or 5 fuzzy sets per fuzzy partition and 10 or 20 maximum rules. Also, we compare the results with three classic crisp rule-based classifiers: decision trees, Ripper-k and PART \cite{Witten2005}. For the data we have used five datasets from the UCI dataset repository \cite{Dua2019}: \textit{appendicitis}, \textit{iris}, \textit{magic}, \textit{vehicle}, \textit{australian}. The reader can find the results in Table \ref{table:datasets}. 

We can notice that the majority of the complexity measures are bellow 0.5 which means that the number of crisp rules could have been more than the double in those situations. Also, although that in some cases the number of fuzzy rules is the same, the number of crisp rules differs a lot. For instance, the number of crisp rules is much lower in the dataset iris than in the dataset australian in the configuration 3-10, even though each fuzzy rule base consists in 5 rules. In other cases the number of fuzzy rules may be higher, but the number of crisp rules has the opposite behavior. For instance, for the dataset australian the configuration 3-10 gives more crisp rules than the configuration 3-20. Now, if we compare with the crisp rule-based algorithms we can notice that the number of crisp rules is usually higher for the fuzzy classifiers. This was to be expected, since fuzzy classifiers are not optimized to minimize the underlying crisp partitions. Finally, we can observe that if we consider only crisp rules defined on disjoint hyperrectangles the number of rules is significantly high.

\section{Conclusions}\label{section:conclusions}
In this work, we have proposed a new algorithm that converts a fuzzy rule-based classifier into a crisp rule-based classifier. We have studied the theoretical upper bounds of complexity that can result in the process and how it behaves in some applied cases. We have also compared the resulting number of rules with other crisp rule-based classifiers.

We believe that our research can lead to improvements in the generation of fuzzy rule-based classifiers, as it opens the possibility of also using crisp rule-based techniques to refine fuzzy classifiers. It also discloses several other research questions, like how to generate a fuzzy classifier with the least added complexity with respect to its simpler crisp equivalent or how to do the reverse process and obtain the simplest fuzzy classifier possible. 

Our future research shall study these challenges in order to find solutions feasible for the practical use of fuzzy systems.

\section*{Acknowledgment}

Raquel Fernandez-Peralta is funded by the EU NextGenerationEU through the Recovery and Resilience Plan
for Slovakia under the project No. 09I03-03-V04- 00557.

Javier Fumanal-Idocin research has been supported by the European Union and the University of Essex under a Marie Sklodowska-Curie YUFE4 postdoc action.

\bibliographystyle{IEEEtran}
\bibliography{bib}

\end{document}